\documentclass{article}

\usepackage{spconf, amsmath, graphicx}
\usepackage{times}
\usepackage{epsfig}
\usepackage{amssymb}
\usepackage{qiangstyle}
\usepackage{multirow}
\usepackage{enumitem}
\usepackage{subcaption}

\usepackage[ruled,vlined]{algorithm2e}

\usepackage{url}
\usepackage[capitalise]{cleveref}
\usepackage{caption,lipsum}
\usepackage{booktabs}
\usepackage{placeins}
\usepackage{color}
\usepackage{enumerate}   
\usepackage[normalem]{ulem}

\usepackage{amsthm}
\newtheorem{prop}{Proposition}

\title{LiCo-Net: Linearized Convolution Network for Hardware-efficient Keyword Spotting}
\name{\parbox{0.98\linewidth}{\centering {Haichuan Yang$^*$\thanks{$^*$Equal Contribution}, Zhaojun Yang$^*$, Li Wan, Biqiao Zhang, Yangyang Shi, Yiteng Huang, Ivaylo Enchev, Limin Tang, Raziel Alvarez, Ming Sun, Xin Lei, Raghuraman Krishnamoorthi, Vikas Chandra}}}

\address{Meta AI}

\begin{document}

%
\maketitle
\begin{abstract}

This paper proposes a hardware-efficient architecture, Linearized Convolution Network (LiCo-Net) for keyword spotting. It is optimized specifically for low-power processor units like microcontrollers. ML operators exhibit heterogeneous efficiency profiles on power-efficient hardware. Given the exact theoretical computation cost, int8 operators are more computation-effective than float operators, and linear layers are often more efficient than other layers. The proposed LiCo-Net is a dual-phase system that uses the efficient int$8$ linear operators at the inference phase and applies streaming convolutions at the training phase to maintain a high model capacity.
The experimental results show that LiCo-Net outperforms single-value decomposition filter (SVDF) on hardware efficiency with on-par detection performance. Compared to SVDF, LiCo-Net reduces cycles by $40\%$ on HiFi4 DSP.

\end{abstract}

\begin{keywords}
Keyword Spotting, Hardware-aware Model Optimization, Power-efficient ML
\end{keywords}

\section{Introduction}
\label{sec:intro}
Voice assistant has become a trendy way for users to interact with smart devices. Keyword spotting (KWS), an essential voice assistant component, continuously runs on the device and detects a predefined keyword. The keyword detection activates the device such that the assistant starts to listen to the voice command and takes the corresponding action. As the gate to voice assistant, KWS saves computation and power by avoiding always running the entire speech
recognition system in the background. In addition, the detection accuracy of KWS is crucial for the perceived user experience of the device, i.e., correctly triggering when the user speaks the keyword, as a failure to wake up on a trigger attempt is frustrating. Since the KWS system is always-on, for devices that have limited battery life and memory space, it is highly desirable to develop an efficient KWS model in the
aspects of computation and memory consumption.

A variety of neural network models have been proposed for KWS modeling in the deep learning era \cite{lopez2021deep}, such as MLPs (fully connected networks)~\cite{chen2014small,zhang2017hello}, convolutional neural networks (CNNs)~\cite{zhang2017hello,sainath15b_interspeech,kim2021broadcasted,tang2018deep,sun2017compressed}, recurrent networks (RNNs)~\cite{fernandez2007application,arik2017convolutional,sun2016max},  and attention models \cite{berg2021keyword,shan2018attention}. 
They have different pros and cons. MLPs have simple model architectures and efficient hardware support. However, due to the restricted model capacity, MLPs are often associated with a large input size and limited accuracy. RNNs show improved performance, but their dependency on the previous states is non-determined, which leads to unstable prediction for the always-on streaming data. Moreover, the nonlinear operators in RNNs are inefficient on low-power hardware and vulnerable to static int8 quantization. In contrast, CNNs are friendly to int8 quantization with high model performance, but the computation cost is usually too high for always-on KWS models.

Recently, researchers have been interested in optimizing CNN architectures to develop accurate and efficient on-device KWS models.
\cite{coucke2019efficient,rybakov20_interspeech} propose to run inference of CNNs in a streaming manner, such that the model output at each timestamp is inferred based on a small new input and the previously computed history. 
The computation cost of CNNs hence can be significantly reduced.
Following the streaming fashion, singular value decomposition filter (SVDF)~\cite{alvarez2019end} and S$1$DCNN~\cite{higuchi2020stacked} further improve the efficiency of CNNs by using depthwise and $1$D convolutions. These model have demonstrated tempting properties of maintaining high detection quality with small-footprint model size and computation.

Although these optimized CNNs are efficient in theoretical computation metrics, e.g., the multiply-accumulates (MACs), such theoretical numbers often cannot be interpreted as real hardware efficiency. One main reason is that depending on the hardware design and specifications, the hardware efficiency of convolutions is often below
expectation. Linear operators, nevertheless, are usually highly optimized on processors so that they can fully utilize the computational capacity. As a result, the hardware efficiency of linear operators could even surpass the number of MACs in some cases. The optimization of linear operators leads to an interesting phenomenon: \emph{an MLP with a higher number of MACs could be more hardware-efficient than CNNs with fewer MACs}.


This paper proposes a novel neural network architecture for hardware-efficient KWS. We carefully design the model architecture to achieve high hardware efficiency and competitive model performance. Inspired by the observation mentioned above on the hardware-efficiency of linear operators and the modeling-efficacy of convolutions, we propose Linearized Convolution Network (LiCo-Net) by pushing the hardware utilization to an optimal case. It is a dual-phase system that uses the computation-effective $8$-bit linear operators at the inference phase and applies streaming convolutions at the training phase to maintain a high model capacity. To summarize the contributions of this paper,
\begin{itemize}[leftmargin=*,noitemsep,topsep=0pt,parsep=0pt,partopsep=0pt]
    \item we demonstrate how to equivalently convert the CNNs to streaming models and further to MLPs, as well as the mathematical requirement for such conversion;
    \item we further propose the basic dual-phase building block LiCo-Block designed as a bottleneck structure of three $1$D convolutions for training which are then equivalently linearized for inference;
    \item we design LiCo-Net by stacking the LiCo-Blocks, showing that it is more hardware-efficient meanwhile achieves on-par model performance compared to the competitive SVDF.
\end{itemize}

\section{Linearized Convolution Network}\label{sec:lico}
CNNs have brought success to the KWS task. A CNN model takes acoustics features as input, e.g., LMELs or MFCCs, and outputs keyword labels, e.g., phonemes or sub-words in the keyword. 
In addition to the extensive work on applying $2$D convolutions for KWS modeling \cite{zhang2017hello,sainath15b_interspeech}, $1$D convolutions have also gained popularity due to the better trade-off between efficiency and effectiveness \cite{choi2019temporal, li2020small}. Furthermore, streaming convolutions have become a common technique for reducing computation cost \cite{rybakov20_interspeech, alvarez2019end}. A streaming convolution infers the output at each timestamp based on a small new input from the streaming audio and the previously computed history. We design the linearized convolution architecture based on the streaming $1$D convolutions in this work. 

\subsection{Formulation of Streaming Convolution}



Given an input $X\in R^{C\times T}$ with $C$ features and $T$ time frames, the output of the $1$D convolution $W\in R^{D \times C \times K}$ with stride $s$
can be computed as $Y = conv(X, W, s)$,
\begin{gather}
 Y_{d,i} = \sum_{c=0}^{C-1} \sum_{k=0}^{K-1} W_{d,c,k} X_{c, s \times i + k}, \label{eq:conv}\\
 \forall i = 0,1,2,..., \lfloor (T-K)/s \rfloor, \ \forall d = 0,1,2,...,D-1, \notag
\end{gather}
where $K$ is the kernel size, $D$ denotes the number of output channels, and $Y \in R^{D\times \lfloor (T-K)/s + 1\rfloor}$.
Without losing generality, we hide the bias terms in Eq.~\eqref{eq:conv}.
\begin{figure}[t]
\centering
\includegraphics[width=0.95\linewidth]{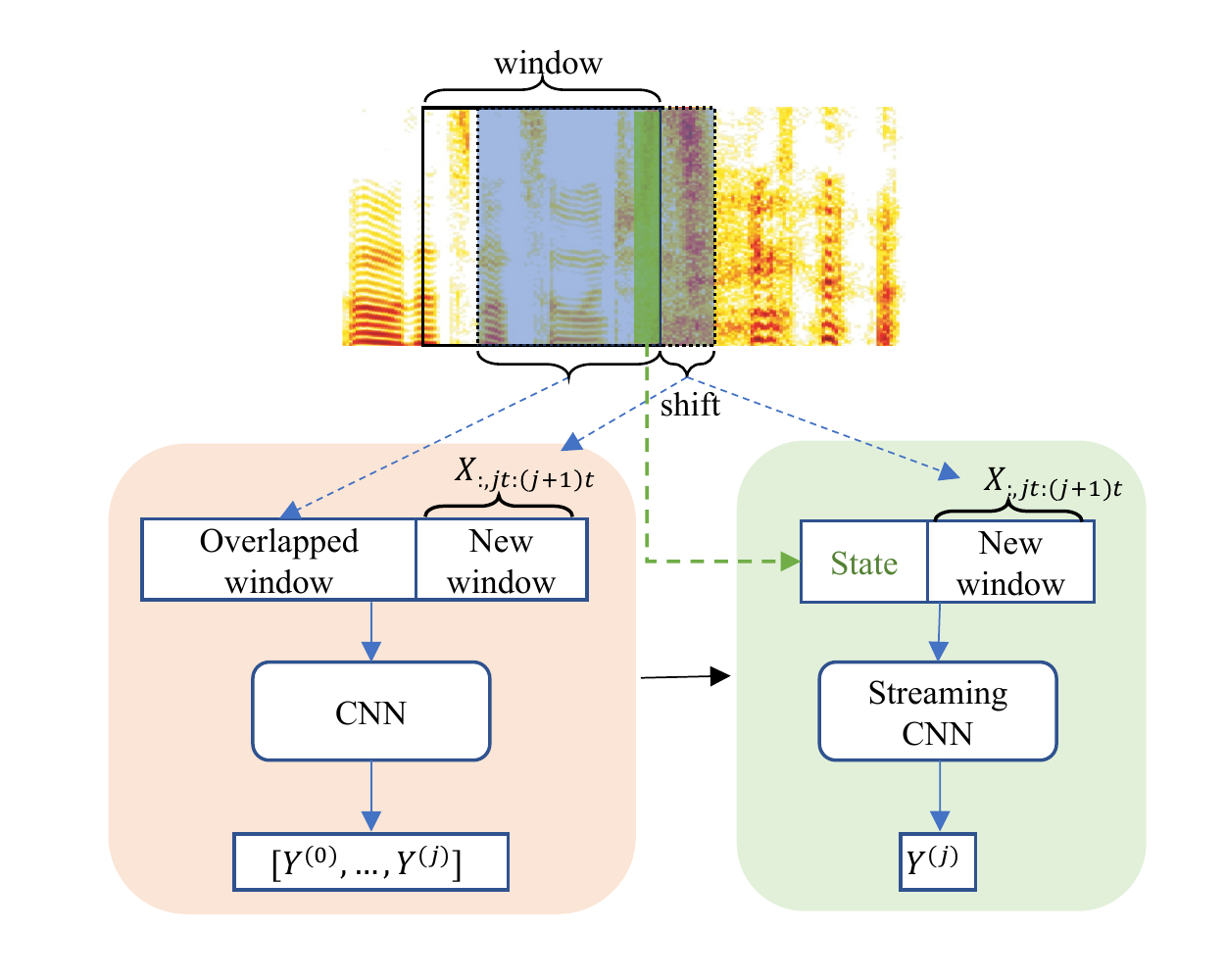}
\vspace{-0.5cm}
\caption{Illustration of streaming CNNs.\vspace{-0.5cm}}
\label{fig:scnn}
\end{figure}
In real-time KWS applications, an audio stream sequentially flows into the system. Suppose that the input $X$ of $T$ frames consisting of $N$ chunks is incrementally observed,\vspace{-0.1cm}
\begin{equation*}
X = [X_{:, 0:t},..., X_{:, (j-1)t:jt},...,X_{:, (N-1)t: T}],
\vspace{-0.1cm}
\end{equation*}
where $\forall j = 1,2,...,N$, and $X_{:, jt:(j+1)t}$ is the $j$-th chunk of size $t$. 
Given a causal convolution kernel,
we can incrementally compute $Y$ as each chunk of $X$ is observed along the time dimension.
If we set $t$ as a multiple of the convolution stride $s$, we can get the same number $\frac{t}{s}$ of output elements for each chunk.
To mathematically align the output as the $j$-th chunk is observed with that computed in the non-streaming manner, 
we pad the $j$-th chunk with $K-s$ length of the most recent history $X_{:,jt-(K-s):jt}$. The partial output $Y^{(j)}$ can hence be inferred as:
\begin{equation}
Y^{(j)} = conv(X^{(j)}, W, s), \label{eq:sconv}
\vspace{-0.1cm}
\end{equation}
where $X^{(j)}=[X_{:,jt-(K-s):jt},\ X_{:,jt:(j+1)t}]$, i.e., the concatenation of the history and the $j$-th chunk, and $Y^{(j)} \in R^{D\times t/s}$. Therefore, the non-streaming output $Y$ can be expressed as the concatenation of $\{Y^{(j)}\}_{j=0}^{J-1}$,
\begin{equation}
Y = [Y^{(0)}, Y^{(1)}, ..., Y^{(J-1)}].
\vspace{-0.1cm}
\end{equation}
We maintain the padding content $X_{:,jt-(K-s):jt}$ as an internal state in the streaming convolution implementation.

\paragraph*{Streaming convolution for KWS modeling.} Fig.~\ref{fig:scnn} illustrates streaming and non-streaming convolutions in the KWS application. At runtime, the KWS model predicts keyword labels over a sliding $window$ with a window $shift$ along the audio stream. In the non-streaming scenario, between two consecutive input windows, there is often a significant overlap $widnow-shift$ that repeatedly participates in computation. In contrast, the streaming convolution equivalently produces the corresponding output using a much smaller input including a small new content $X_{:,jt:(j+1)t}$ and a maintained internal state $X_{:,jt-(K-s):jt}$, resulting in a significant computation reduction.

\subsection{Linearized Convolution Block}
We notice that a conventional $1$D convolution is equivalent to a linear layer, if the input has the same length as the kernel size. Therefore, Eq.~\eqref{eq:sconv} can be reformulated as a linear layer, if $X^{(j)}\in R^{C \times K}$, 
\begin{equation}
Y^{(j)}=linear(\tilde{X}^{(j)}, \tilde{W}),
\label{eq:linear}
\vspace{-0.2cm}
\end{equation}
by reshaping the convolution parameters $W\in R^{D \times C \times K}$ to $\tilde{W}\in R^{CK \times D}$ as well as $X^{(j)} \in R^{C \times K}$ to $\tilde{X}^{(j)} \in R^{1 \times CK}$. Interestingly, in the streaming scenario, when the chunk size $t$ equals the stride $s$, the  $X^{(j)}$ length also equals $K$.
This observation leads to the condition for the equivalent reformulation of a streaming convolution as a linear layer:
\emph{the streaming chunk size $t$ equals the convolution stride $s$}.
In real-time applications, the stride $s$ implies how frequent the KWS model performs an inference. A larger $s$ indicates a lower inference frequency, resulting in a lower computation cost but potentially a higher latency. To generalize the linearization to a streaming CNN, we introduce the proposition below.

\begin{figure}[t]
\centering
\includegraphics[width=0.9\linewidth]{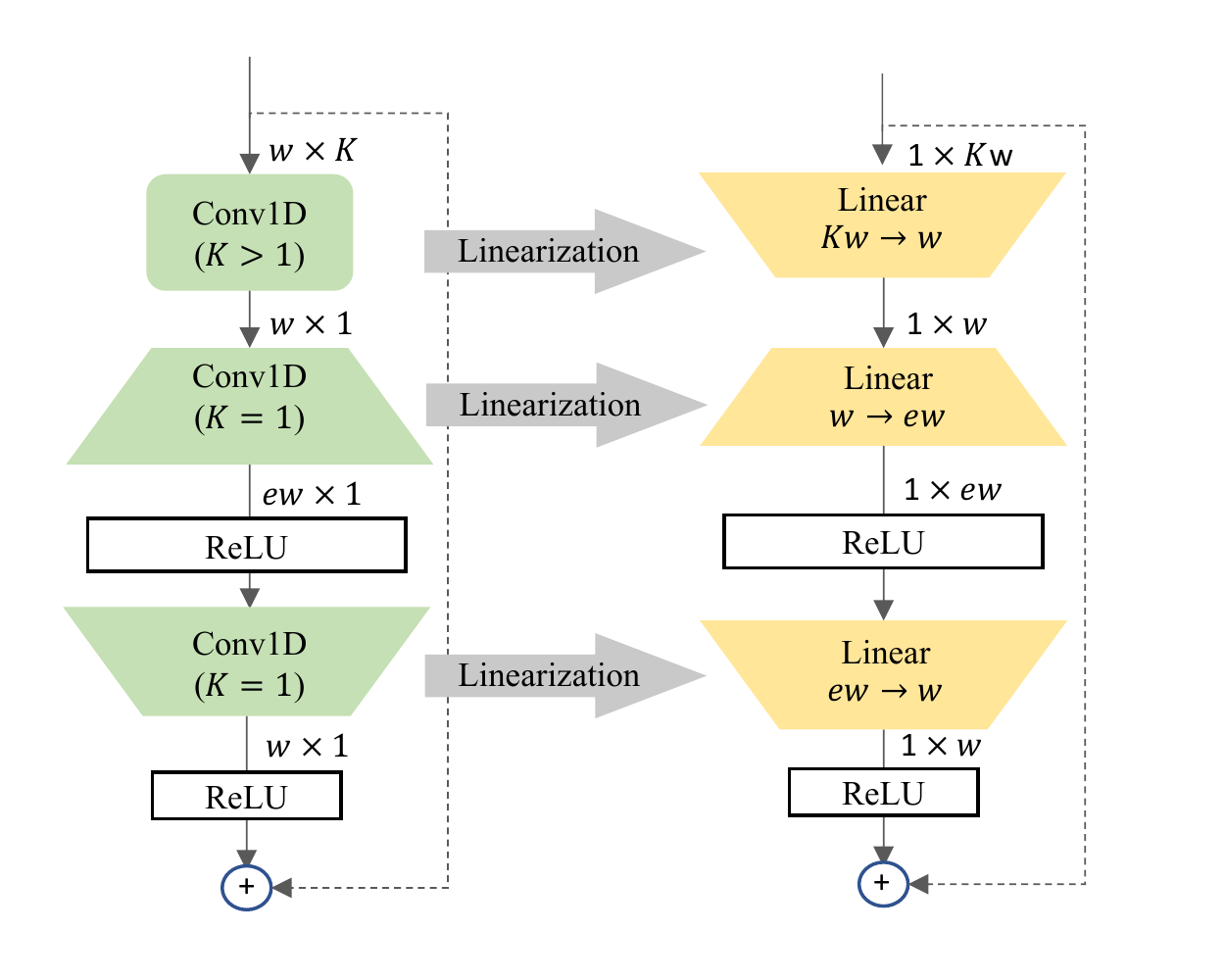}
\vspace{-0.6cm}
\caption{Linearized Convolution Block. \vspace{-0.6cm}
}
\label{fig:lico}
\end{figure}

\begin{prop}\label{prop1}
A streaming CNN can be equivalently converted to an MLP if it meets the following conditions:
\begin{itemize}[leftmargin=*,noitemsep,topsep=0pt,parsep=0pt,partopsep=0pt]
    \item The first $1$D convolution layer has equal stride $s$ and chunk size $t$;
    \item All the other layers have strides equal to 1.
\end{itemize}
\vspace{-0.1cm}
\end{prop}
We hence propose the Linearized Convolution (LiCo) block with careful engineering design to optimize the model capacity and the number of parameters. 
Fig.~\ref{fig:lico} illustrates the architecture by using regular $1$D convolutions for simplicity.
LiCo-Block consists of three $1$D convolution layers: the first layer has a kernel size greater than $1$, followed by two point-wise convolutions. We enforce a bottleneck structure between the layers to further boost modeling effectiveness. Within the block, the inner size of the expanded representation is controlled by the expansion factor $e > 1$. A residual connection is also introduced when $s=1$ for effective gradient propagation. Given an input with a length $K$, each convolution layer can be equivalently linearized.


We further construct LiCo-Net by stacking $L$ LiCo-Blocks in sequence. Following Proposition~\ref{prop1}, we use $s_1 \geq 1$ for the first building block and $s_l = 1$ for the rest of the blocks.
We adopt streaming convolutions in practice. Hence we also enforce $t = s_1$ to linearize the convolutions equivalently. LiCo-Net is a dual-phase system that applies streaming convolutions at the training phase which are then linearized for efficient inference.

\section{Experiments}\label{sec:exp}
%
%
\subsection{Dataset}
We choose the generic phrase ``hey operator'' as the keyword in the experiments. The aggregated and de-identified dataset contains $95$K utterances collected through crowd-sourced workers.
We split the data into training and testing sets in a speaker-independent way. The training set contains $84,354$ utterances from $10,047$ speakers and the testing set has $10,765$ utterances from $1,202$ speakers. We use irrelevant speech from various languages without the keyword as the negative data. The training and testing sets contain $306$ and $194$ hours of negative data, respectively.
We further augment the training data with background noise and speed perturbation to increase data variability.
We extract acoustic features using $40$-dimensional log Mel-filterbank energies computed over a $25$ms window every $10$ms. Finally, global normalization is applied to the features. 
\begin{figure*}[t]
     \centering
     \begin{subfigure}[b]{0.49\linewidth}
         \centering
         \includegraphics[width=7.8cm]{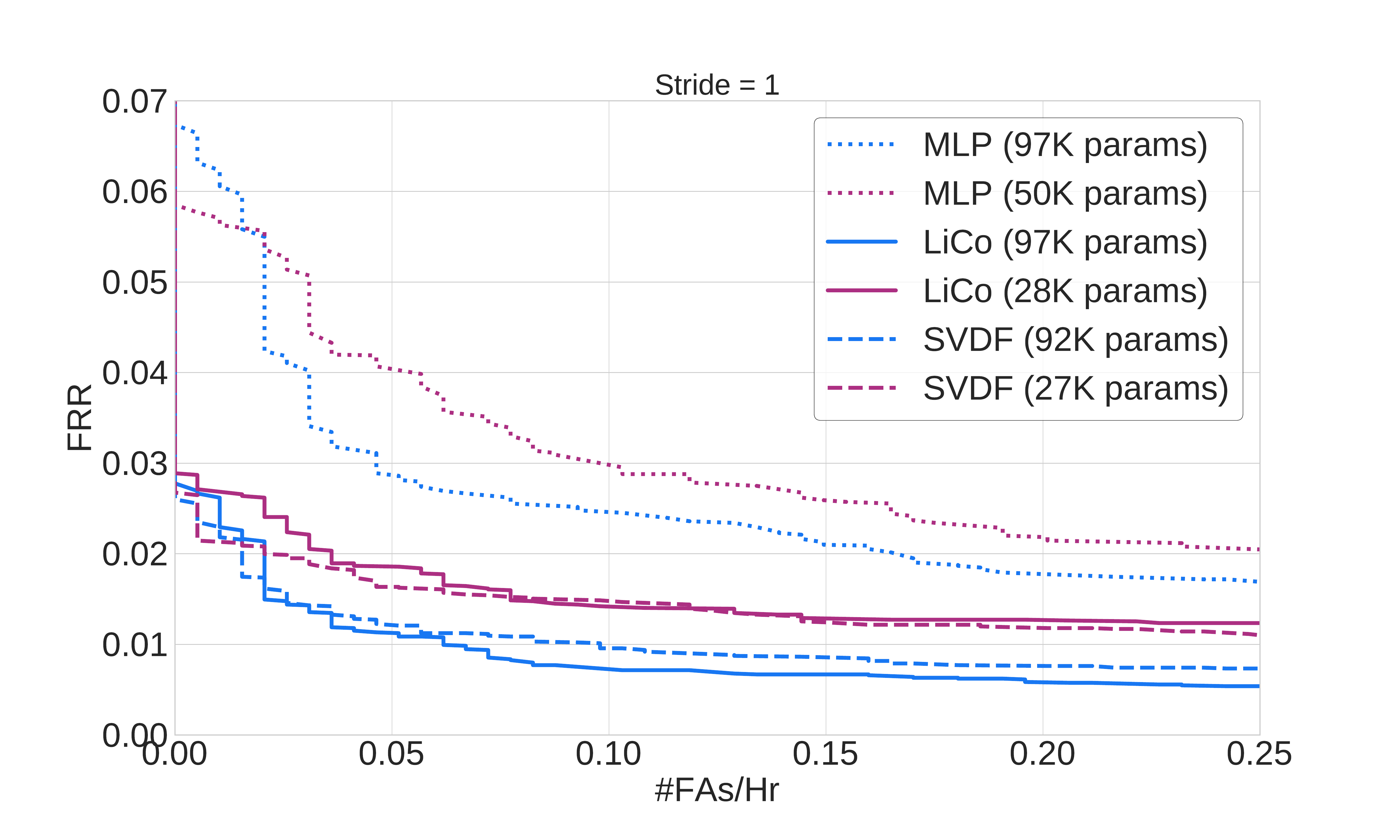}
         \vspace{-0.3cm}
     \end{subfigure}
     \begin{subfigure}[b]{0.49\linewidth}
         \centering
         \includegraphics[width=7.8cm]{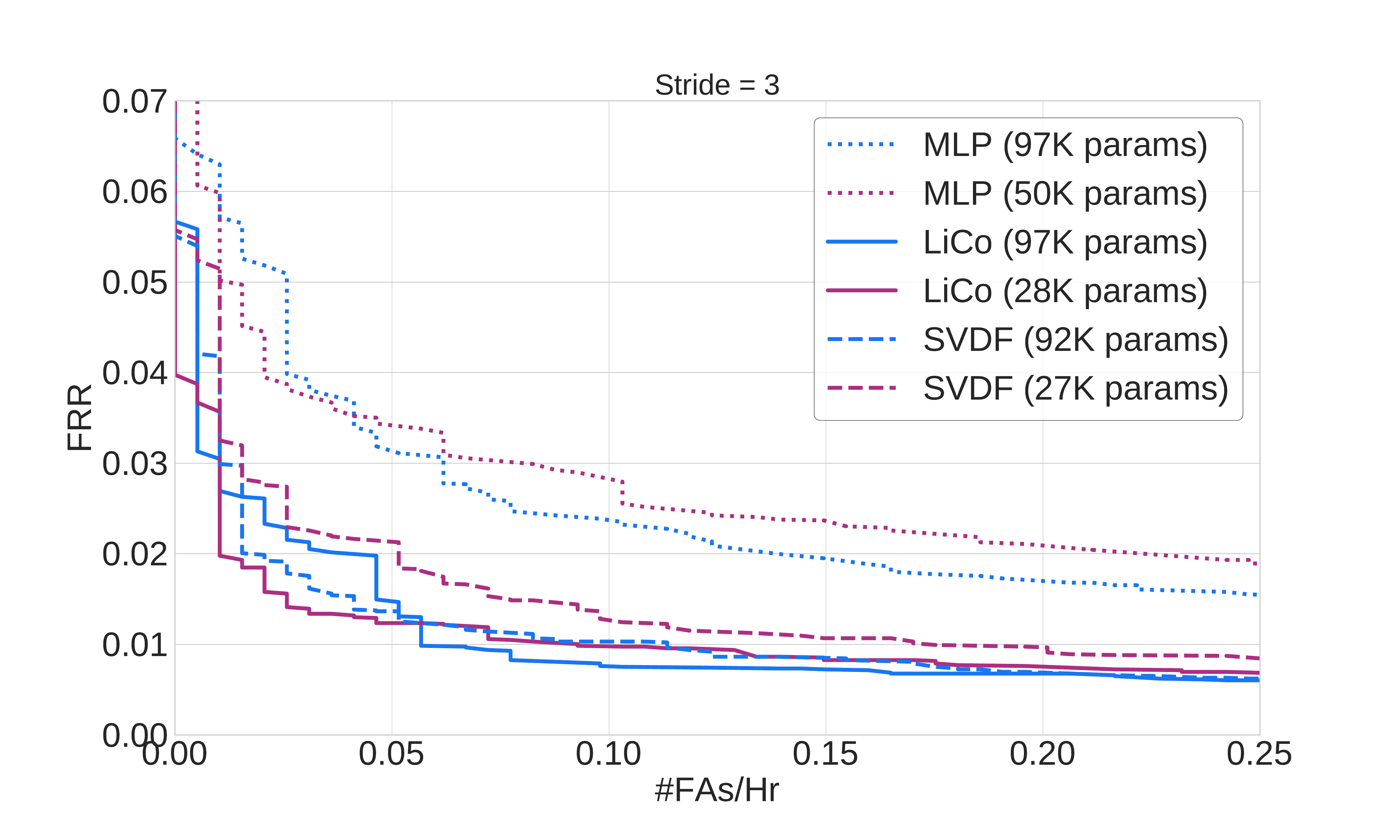}
         \vspace{-0.3cm}
     \end{subfigure}
\caption{DET curves of different models with $\textrm{stride}=1$ (left) and $\textrm{stride}=3$ (right).\vspace{-0.5cm}}
\label{fig:det}
\end{figure*}
\subsection{Experimental Setup}
{\bf Model architectures.} We conduct the experiments on $3$ model types: LiCo-Net, SVDF, and MLP. We design both small and large versions for each model type. We construct LiCo-Net by stacking 5 LiCo-Blocks. Specifically we set expansion factor $e=6$, kernel size $K=5$, and channel width $w=32$ for the large model ($96.6$K), and $e=4, K=4, w=16$ for the small model ($28.4$K). 
We also construct SVDF using $5$ blocks, each of which contains one SVDF layer and a bottleneck layer~\cite{alvarez2019end}. The large SVDF uses kernel size $K=5$, $256$ filters, and a bottleneck size of $32$ ($92$K). The small SVDF uses kernel size $K=4$, $128$ filters, and a bottleneck size of $16$ ($26.9$K).
The MLP models take the audio sequence of $21$ frames as input and consist of two hidden layers. It is equivalent to using 1 LiCo-Block with kernel size $=21$. The large MLP uses $80$ neurons in the first layer and $320$ neurons in the second layer ($96.7$K). The small MLP uses $40$ neurons in the first layer and $320$ neurons in the second layer ($50.3$K). In each version, the model sizes of different model types are comparable. We experiment with each model using $s=1$ and $s=3$ in the first layer to investigate stride's impact on model performance and efficiency. All the models have a linear classifier at the end.

{\bf Training and evaluation protocols.} KWS models are trained to predict $11$ targets, including $9$ subwords in the keyword phrase, SIL, and FILLER. We obtain the target labels through force alignment using a pre-trained acoustic model.
We use a batch size of $256$ with $8$ GPUs for training and Adam optimizer with a learning rate of $0.01$ and $\beta=(0.9, 0.98)$.
At the evaluation stage, the model inference is executed in a streaming manner, i.e., a model performs on $s$ feature frames at the step of $s$, following proposition \ref{prop1}. Therefore, a larger stride suggests fewer inferences per second.
An external decoder similar to \cite{prabhavalkar2015automatic} is used to aggregate the smoothed frame-wise posteriors within a $1.1$s window into a final detection score ranging from $0$ to $1$.
We present model performance by plotting detection error trade-off (DET) curves, where the x-axis and y-axis represent the number of false accepts (FA) per hour and false reject rate (FRR), respectively. In order to measure hardware efficiency, all the models are $8$bits quantized and are profiled on a system with Cadence Tensilica HiFi4 DSP core~\cite{rt600mcu}, using the toolchain provided cycle-accurate instruction set simulator. We use DSP cycles per inference and million instructions per second (MIPS) as the metrics.
For all the models, we use the optimized kernel for their operators, e.g., SIMD instructions for implementing depthwise convolution.




\section{Results and Discussions}\label{sec:results}
\subsection{Comparison of Model Performance}
Fig.~\ref{fig:det} presents the DET curves of different models with $\textrm{stride}=1$ and $\textrm{stride}=3$. We can observe that the large model of each model type exhibits better performance compared to the corresponding miniature version. When $\textrm{stride}=3$, the gap between the two LiCo-Nets is much less phenomenal. SVDF and LiCo-Net have achieved similarly promising performance in most cases. The small LiCo-Net outperforms the small SVDF when $\textrm{stride}=3$ by improving FRR from $2\%$ to $1.2\%$ at $0.025$ FAs/Hr. This observation indicates that LiCo-Net might have strong modeling power even with a small number of parameters. The performance of both SVDF and LiCo-Net significantly surpasses MLP. It is also interesting to see that a big stride barely affects the performance of all the models. This is a tempting property since a more significant stride guarantees better computation effectiveness without sacrificing model performance.

\begin{table}[htbp]
\fontsize{8}{10}\selectfont
\centering
\caption{Profiling results of different models on HiFi4 DSP.}
\vspace{-0.1cm}
\label{table:profile_res}
\begin{tabular}{c|c|cccc}
\toprule[2pt]
                         & Model     & Params   & MACs & Cycles & MIPS \\
\toprule[2pt]
\multirow{3}{*}{\begin{tabular}[c]{@{}c@{}}$\textrm{stride}=1$\\ Small model\end{tabular}} & MLP    & $50.3$K     & $49.9$K       & $34,499$     &  $3.4$   \\
                          & SVDF       &  $26.9$K    &  $26.3$K      &  $90,751$    &   $9.07$     \\
                          & LiCo-Net   &  $28.4$K    &  $27.8$K      &  $60,808$    &   $6.08$     \\
                          \toprule[0.5pt]
\multirow{3}{*}{\begin{tabular}[c]{@{}c@{}}$\textrm{stride}=1$\\ Large model\end{tabular}}  & MLP   &  $96.7$K    &   $96.3$K    &   $48,891$   &   $4.88$     \\  
                          & SVDF     &  $92$K    &  $90.7$K      &   $152,458$   &    $15.2$    \\
                          & LiCo-Net &   $96.6$K   &   $95.3$K     &   $93,622$   &   $9.36$     \\
                          \toprule[1.5pt]
\multirow{3}{*}{\begin{tabular}[c]{@{}c@{}}$\textrm{stride}=3$\\ Small model\end{tabular}} & MLP         &   $50.3$K   &  $49.9$K      & $35,119$     &  $1.15$      \\
                          & SVDF     &   $26.9$K   &  $36.5$K      &   $100,946$   &   $3.33$     \\
                          & LiCo-Net & $28.4$K     &   $27.8$K     &  $61,154$    &   $2.01$     \\
                          \toprule[0.5pt]
\multirow{3}{*}{\begin{tabular}[c]{@{}c@{}}$\textrm{stride}=3$\\ Large model\end{tabular}}  & MLP      &  $96.7$K    &   $96.3$K     &  $49,258$    &   $1.62$     \\
                          & SVDF     &  $92$K    &   $111.2$K     &  $169,922$    &  $5.6$      \\
                          & LiCo-Net &  $96.6$K    &   $95.3$K     &  $94,556$    &  $3.12$     \\
\toprule[2pt]
\end{tabular}
\vspace{-0.2cm}
\end{table}

\subsection{Improvement on DSP Efficiency}
\vspace{-0.1cm}
Table 1 summarizes the profiling results of different models.
We can see that among the models with similar MACs and
stride, LiCo-Net is $1.5\sim1.8$x faster on DSP than SVDF, while costs $2$x more DSP cycles compared to MLP. Though the model size of the small MLP is almost twice as big as the other two opponents, it still exhibits the highest efficiency. Compared to LiCo-Net, MLP contains much fewer operators, especially linear operators, so it consumes fewer DSP cycles even with higher MACs. This observation has demonstrated the efficiency superiority of linear operators on hardware. Note that the efficiency of operator implementation highly depends on the hardware design and specifications. The gap between SVDF and LiCo-Net on HiFi4 DSP could be different on another chip.


\label{sec:ablation}
\section{Conclusion}
\vspace{-0.3cm}
In this work, we propose Linearized Convolution Network (LiCo-Net), a hardware-efficient network for KWS. It is a dual-phase system that uses the computation-effective $8$-bit linear operators at the inference phase and applies streaming convolutions at the training phase to maintain a high model capacity.
As a general efficient architecture, LiCo-Net can also be applied for other always-on speech tasks with stringent latency and power consumption budgets.

\bibliographystyle{IEEEbib}
\bibliography{reference}

\end{document}